\pdfoutput=1

\documentclass[11pt]{article}

\usepackage[preprint]{acl}

\usepackage{times}
\usepackage{latexsym}
\usepackage[T1]{fontenc}

\usepackage[utf8]{inputenc}
\usepackage{microtype}

\usepackage{inconsolata}
\usepackage{hyperref}
\usepackage{url}
\usepackage{xcolor}  
\usepackage{tcolorbox}
\usepackage{booktabs} 
\usepackage{multirow} 
\usepackage{graphicx}
\usepackage{pifont} 
\usepackage{hyperref}
\usepackage{arydshln}
\usepackage{listings}
\usepackage{algorithm}
\usepackage{amsmath} 
\usepackage{algpseudocode}
\definecolor{codegreen}{rgb}{0,0.6,0}
\definecolor{codegray}{rgb}{0.5,0.5,0.5}
\definecolor{codepink}{RGB}{252, 142, 172}
\definecolor{codepurple}{rgb}{0.58,0,0.82}
\definecolor{backcolour}{RGB}{245,245,245}
\lstdefinestyle{mystyle}{
    backgroundcolor=\color{backcolour},   
    commentstyle=\color{magenta},
    keywordstyle=\color{blue},
    numberstyle=\tiny\color{codegray},
    stringstyle=\color{codepurple},
    basicstyle=\fontfamily{\ttdefault}\footnotesize,
    breakatwhitespace=false,         
    breaklines=true,                 
    keepspaces=true,    
    frame=single,
    numbersep=1pt,                  
    showspaces=false,                
    showstringspaces=false,
    showtabs=false,                  
    tabsize=1,
    classoffset=1, 
    keywordstyle=\color{violet},
    classoffset=0
}
\lstdefinelanguage{JavaScript}{
  keywords={typeof, new, true, false, catch, function, return, null, catch, switch, var, if, in, while, do, else, case, break},
  keywordstyle=\color{blue}\bfseries,
  ndkeywords={class, export, boolean, throw, implements, import, this},
  ndkeywordstyle=\color{darkgray}\bfseries,
  identifierstyle=\color{black},
  sensitive=false,
  comment=[l]{//},
  morecomment=[s]{/*}{*/},
  commentstyle=\color{purple}\ttfamily,
  stringstyle=\color{red}\ttfamily,
  morestring=[b]',
  morestring=[b]"
}

\usepackage{devanagari}
\title{IntentGPT: Few-shot Intent Discovery with Large Language Models}

\author{Juan A. Rodriguez \\
  ServiceNow Research\\
  Mila - Quebec AI Institute \\
  École de Technologie Supérieure \\
   \\\And
  Nicholas Botzer \\
  ServiceNow Research \\
  University of Notre Dame \\\And
  David Vazquez \\
  ServiceNow Research \\\AND
  Christopher Pal \\
  ServiceNow Research\\
  Mila - Quebec AI Institute \\\And
  Marco Pedersoli \\
  École de Technologie Supérieure \\
  ServiceNow Research\\\And
  Issam Laradji \\
  ServiceNow Research 
}

\begin{document}
\maketitle

\begin{abstract}
In today's digitally driven world, dialogue systems play a pivotal role in enhancing user interactions, from customer service to virtual assistants. In these dialogues, it is important to identify user's goals automatically to resolve their needs promptly. This has necessitated the integration of models that perform Intent Detection. However, users' intents are diverse and dynamic, making it challenging to maintain a fixed set of predefined intents. As a result, a more practical approach is to develop a model capable of identifying new intents as they emerge. We address the challenge of Intent Discovery, an area that has drawn significant attention in recent research efforts. Existing methods need to train on a substantial amount of data for correctly identifying new intents, demanding significant human effort. To overcome this, we introduce IntentGPT, a novel training-free method that effectively prompts Large Language Models (LLMs) such as GPT-4 to discover new intents with minimal labeled data. IntentGPT comprises an \textit{In-Context Prompt Generator}, which generates informative prompts for In-Context Learning, an \textit{Intent Predictor} for classifying and discovering user intents from utterances, and a \textit{Semantic Few-Shot Sampler} that selects relevant few-shot examples and a set of known intents to be injected into the prompt. Our experiments show that IntentGPT outperforms previous methods that require extensive domain-specific data and fine-tuning, in popular benchmarks, including CLINC and BANKING, among others. 
\end{abstract}

\section{Introduction}\label{introduction}
Intent Discovery is a natural language processing (NLP) task that involves classifying user-written sentences into intents (goals) within an \textit{open-world} context, where new intents will appear and need to be identified. This task is crucial for modern dialogue systems~\citep{degand2020introduction}, allowing them to decipher user queries, whether they involve seeking information, making requests, or expressing opinions, and steering the conversation appropriately. This task is dynamic since user intents evolve and new digital tools continually emerge, expanding the range of intents. Intent prediction needs to adapt to this open-world scenario and stay responsive to changing user needs.

\begin{figure*}[t]
    \centering
    \includegraphics[width=1.0\textwidth, trim={0 35px 0 45px}, clip]{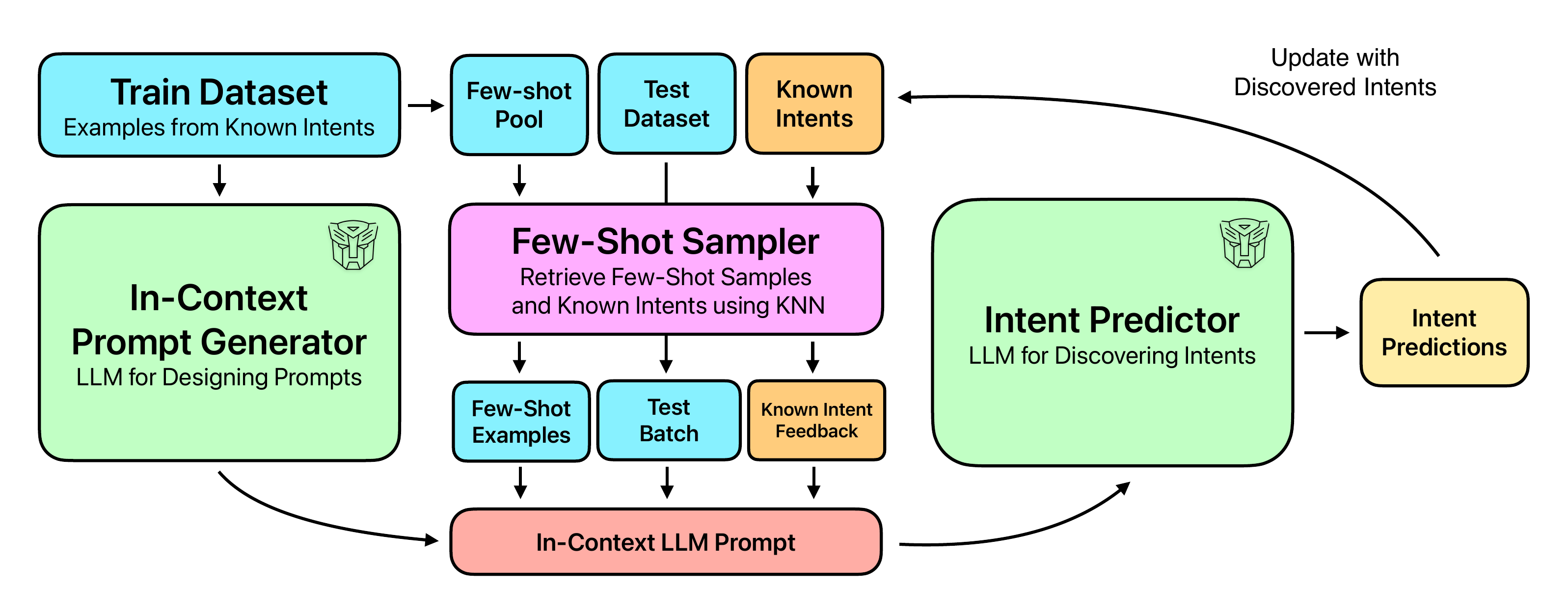}
    \vspace{-4mm}
    \caption{\textbf{IntentGPT architecture.} The system consists of two frozen LLMs. The In-Context Prompt Generator receives a small subset of training data and is devoted to generating a task prompt using domain context. The generated prompt is filled with, few-shot samples, a list of known intents, and the test sample (user query).  Finally, the Intent Predictor assigns (or discovers) the intent and updates the Known Intents database accordingly.} 
    \vspace{-4mm}
    \label{fig:figure1}
\end{figure*}


Intent Detection has received considerable research in recent years~\citep{lin2023selective, costello2018multi,fitzgerald2022massive,jayarao2018intent, zhang2020discriminative}. These works approach the problem from a closed-world perspective, with predefined classes, while often allowing for predictions on out-of-scope examples into an additional "other" class. Our focus lies in Intent Discovery~\citep{shi2018auto,zhang2022new}, which assumes an open-world setting where novel intents must be discovered. This approach aligns better with real-world applications.

The task of Intent Discovery has seen notable exploration through learning powerful text representations and aligning clusters~\citep{kumar2022intent,zhang2021discovering}. However, the effectiveness of these approaches often hinges on the availability of substantial labeled data and multi-stage training procedures. Implementing such methods requires human expert annotations for the challenging task of defining relevant intents.

Given these challenges, using pre-trained large language models (LLM) emerges as a compelling alternative. Pre-trained models, such as GPT 3-4~\citep{brown2020language, openai2023gpt4}, have undergone extensive training on vast and diverse text corpora encompassing a wide range of domains and contexts. This pre-training endows them with a broad understanding of language, making them adept at handling complex and diverse user inputs~\citep{bubeck2023sparks, rodriguez2023starvector}. This offers a new perspective to handling open-set problems. As framed by~\citet{scheirer2012toward}, open-set problems pertain to managing the empirical risk space connected with unknown classes. We can consider a model such as GPT-4 to possess extensive knowledge of the world due to its training data. This suggests that the model should not have a problem with the concept of empirical risk space in an open-world.

With the increasing availability of open-source LLMs\footnote{\url{https://huggingface.co/spaces/HuggingFaceH4/open_llm_leaderboard}}, the community has embraced a new paradigm for adapting these models to downstream tasks by crafting carefully designed prompts~\citep{liu2023pre}. This paradigm has demonstrated that, with meticulous prompt construction, these models can deliver precise predictions. Notably, prompt design has proven remarkably influential in shaping LLM behavior~\citet{brown2020language}. Prompts typically comprise a task description and several canonical examples, constituting a framework for in-context learning (ICL)~\citep{xie2021explanation}. 

In this work, we propose \textbf{IntentGPT}, a training-free method for Few-Shot Intent Discovery using pre-trained LLMs. IntentGPT leverages a few labeled examples for prompt generation and learns new intents during inference by injecting discovered intents back into the prompt. Using automatic prompts and access to a database of intents, IntentGPT offers a flexible solution when selecting known intents or discovering new ones. Our design offers a training-free and model-agnostic solution that relies strictly on model generalization and ICL capabilities. Our method assumes a strong world knowledge of large pre-trained models like GPT-3.5, GPT-4, and Llama-2~\citep{touvron2023llama2}, and leverages them for 1) designing In-Context Prompts and 2) performing Intent Discovery. As shown in Figure~\ref{fig:figure1}, we propose techniques to promote intent reusability and inject semantically meaningful examples into the prompt. 

\textbf{Contributions.} In summary, our contributions are: \textbf{i)} Introduce IntentGPT, a training-free method for Intent Discovery using LLMs. \textbf{ii)} Showcase how LLMs can now perform a tasks that previously required extensive data and training, by using techniques for automatic prompt design and self-improvement (Known Intent Feedback) or semantic few-shot selection. \textbf{iii)} Conduct experiments on challenging Intent Discovery benchmarks, showcasing our method's competitive performance against models demanding more data and training. \textbf{iv)} Provide a comprehensive analysis of hyperparameter influence on metric performance in the novel few-shot ICL setting using frozen LLMs.


\section{Related Work}\label{related-work}


\textbf{Intent Detection.} Intent Detection focuses on classifying pre-defined user intents, typically through text sentences~\citep{liu2019review}. \citet{bunk2020diet} introduced a transformer-based~\citep{vaswani2017attention} architecture that jointly learns entity recognition and Intent Detection. \citet{lin2023selective} detects the lack of annotated data as the primary challenge for this task, and uses LLMs for data augmentation. 


Despite the attention, this scenario lacks realism as it does not account for the emergence of new, out-of-distribution (OOD) intents in the real world.

\vspace{3px}
\textbf{Open-World Intent Detection.} aims to classify known intents and identify OOD examples, assigning them to an unknown class. Previous work by \citet{shu2017doc} on open-world text classification used a CNN with a final 1-vs-rest layer and a sigmoid activation function. Recent methods have used KNN-contrastive learning~\citep{zhou2022knn}, synthetic examples~\citet{zhan2021out}, and adaptive decision boundaries~\citep{zhang2021deep}. Other techniques seek to leverage traditional fine-tuning~\citep{devlin2018bert} paradigms thresholding softmax outputs~\citep{zhang2020discriminative}.

While detecting OOD samples is a significant step toward a real-world scenario, it does not fully address the issue. Ideally, we want our model to proactively generate new intents by design.

\vspace{3px}
\textbf{Intent Discovery.} In the context of Intent Discovery, the model operates under the assumption of an \textit{open-world}, where certain intents are known while others remain undiscovered, continuously advancing in its quest to uncover them. The task presents a challenge as the model must exhibit discriminative capabilities, not only for classifying known examples but also for effectively distinguishing out-of-distribution samples that may be present in the test set. Prior methods involved learning robust text representations and aligning clusters for known and new intents. DeepAligned~\cite{zhang2021discovering} employs a BERT~\citep{devlin2018bert} model to extract intent embeddings and use them to learning cluster representations. SCL~\cite{shen2021semi} adopts a Siamese network to discern whether two utterances align with the same intent. DSSCC~\citep{kumar2022intent} pre-trains a SentenceBert model~\citep{reimers2019sentence} on labeled utterances and intents, and employs contrastive learning on labeled and unlabeled samples to learn effective clusters. Recently, \citet{zhou2023probabilistic} proposed a probabilistic framework treating the intent assignments as latent variables. IDAS~\citep{de2023idas} is the method that is closest to ours. They pose the task as summarization problem, where summaries are generated from utterances and clustered.

While these methods show promising results, their reliance on multiple training stages and labeled data can be suboptimal. Our novel few-shot ICL setting eliminates the need of training and heavily reduces human intervention and train data .

\vspace{3px}
\textbf{In-context Learning.} In-context learning (ICL) studies how to condition LLMs to execute downstream tasks without fine-tuning them. ICL involves prompting a pre-trained LLM with a task description and some problem demonstrations, such as injecting input-label pairs into the prompt (few-shot ICL)~\citep{brown2020language,xie2021explanation}. The application of few-shot ICL has yielded substantial improvements for LLMs across a spectrum of NLP tasks~\citep{liang2022holistic}. A major driver of this shift was the rise of emergent properties of LLMs from scaling them up~\citep{wei2022emergent}. \citet{bubeck2023sparks} study the rising capabilities and implications of largely scaled LLMs, e.g., GPT-4 and showcase their performance on complex downstream tasks like coding or vision.

While many ICL setups utilize only a few labeled examples, ongoing research explores prompt optimization. In this regard, two key challenges emerge: selecting effective prompts and identifying suitable few-shot examples. To address prompt selection, \citet{brown2020language} propose \textit{prompt engineering} involving iterative trial-and-error, albeit with human effort. AutoPrompt~\citep{shin2020autoprompt} tackles this issue through a gradient-based search strategy. Prompt Tuning~\citep{lester2021power} introduces learnable weights for specific tokens while maintaining frozen LLM weights. For identifying optimal few-shot samples, prior approaches have focused on Example Retrieval~\citep{rubin2022learning}. \citet{gao2023ambiguityaware} highlight the effectiveness of combining ICL with semantic similarity methods, outperforming fine-tuning in various downstream tasks. \citet{wang2023voyager} showcased GPT-4's exceptional capabilities in domains like Minecraft, suggesting potential applications in an \textit{open-world}.

Our approach for prompt design involves using strong LLMs, such as GPT-4, prompted with labeled examples to generate a precise prompt that offers domain-specific guidelines and context, reminiscent to what is done with Chain of thought (COT)~\cite{chu2023survey}. For selecting few-shot examples, we employ a retrieval module to extract semantically relevant samples. We present the novel approach of injecting the known intents also into the prompt, giving to the LLM direct access to them. When new intents are discovered, they are also injected into the prompt, allowing the model to learn at test time.

\begin{figure*}[ht]
    \centering
    \includegraphics[width=1.0\textwidth, trim={0 38px 0 50px}, clip]{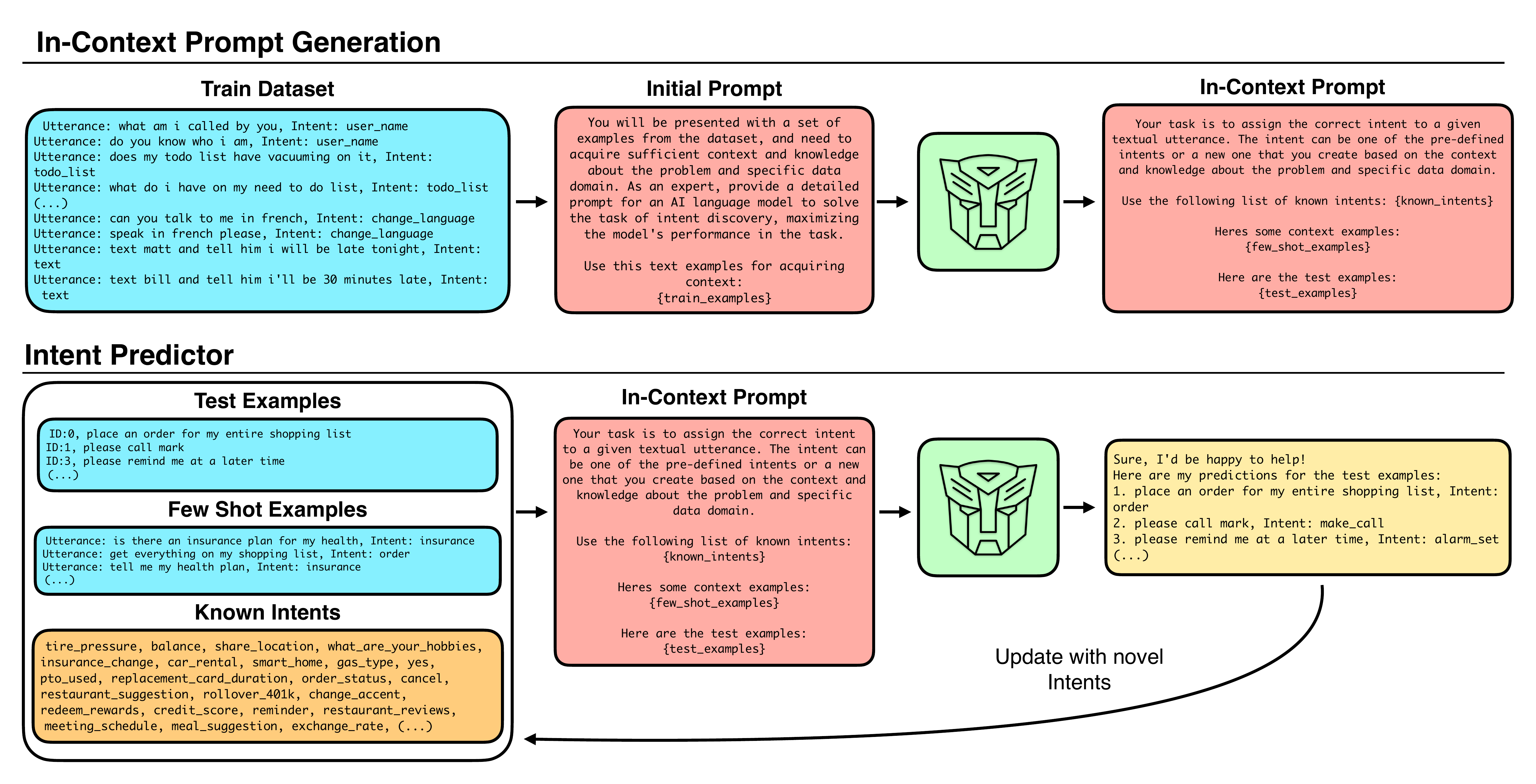}
    \vspace{-4mm}
    \caption{Pipeline of IntentGPT with real examples (CLINC). The top part displays the automatic prompt generation process. The bottom part shows the inference process, where we use the generated prompt and few-shot examples for discovering intents. We do not show the Few Shot Sampler for simplicity.}
    \label{fig:figure2}
    \vspace{-5mm}
\end{figure*}
\section{Method}\label{method}

This section introduces IntentGPT, a training-free and model-agnostic method for Intent Discovery. We propose a novel few-shot ICL setting for Intent Discovery, which uses substantially less data than previous methods. As shown in Figure~\ref{fig:figure1}, IntentGPT comprises two pre-trained transformer LLMs. The \textit{In-Context Prompt Generator} LLM autonomously designs a prompt for the Intent Discovery task, leveraging pairs of utterances and intents from the training set to grasp the domain, contextual information, and language. This LLM is prompted only once, to assist in generating a comprehensive prompt with guidelines, as a form of chain-of-thought\cite{chu2023survey}. On the other hand, during inference, the \textit{Intent Predictor} LLM utilizes the generated prompt, with injected few-shot samples, list of known intents, and query (test) examples to perform inference. Notably, we also include the discovered intents found at test time (\textit{Known Intent Feedback}), allowing IntentGPT to learn on the fly during inference. IntentGPT offers good flexibility between re-using known intents and discovering new ones. Furthermore, the intermediate \textit{Few-Shot Sampler} module, implements various techniques to infuse meaningful information into the prompt. Figure~\ref{fig:figure2} depicts the two stages of our method with real examples.


\subsection{In-Context Prompt Generator}
As ICL performance is substantially influenced by prompt quality, our focus centers on establishing a systematic methodology for crafting effective prompts for the Intent Discovery task. This module is in charge of automatically creating a prompt that explains the task, using small training data for obtaining context, e.g., general domain, such as finance, travel, daily needs, or the preferred language. This eliminates the need for a human expert to manually craft the prompt. We propose the utilization of a robust LLM, such as GPT-4, for prompt design. Specifically, we use examples from the training dataset to give context and ask the In-Context Prompt Generator (ICPG) to solve the task of generating a prompt for Intent Discovery. The ICPG offers intermediate reasoning with guidelines, similar to what is done in Chain-of-Thought (CoT)\cite{chu2023survey}. This stage is only executed once. We show that this improves the quality of the predictions through ablation studies in Section~\ref{ablation}. This module is also used for supporting data in different languages, as the prompt will also be translated (see Appendix~\ref{multilingual} for an analysis on multilingual data).

We initiate the prompt design process by first engaging with the LLM (GPT-4 in our experiments) to ensure the correct output format. Subsequently, we choose a random selection of $x$ examples per known intent ($x=2$ in our experiments) and integrate them into the prompt. This step is essential to provide sufficient context regarding the task and the data domain. By conditioning the LLM on a substantial amount of data, we enable it to grasp the intricacies of both the data and context, thereby facilitating the design of a prompt for another LLM tasked with solving the Intent Discovery task. In the ICPG prompt, we emphasize that the prompt must be concise and explicitly instruct that the 'unknown' intent should never be assigned. For the complete prompt at this stage, please refer to Appendix~\ref{prompt-design} (Prompt A). This process is a one-time operation for a given number of known intents $n$ and the desired number of examples per intent $x$. Once generated, the prompt is retained for future use within the same setting. For examples of prompts generated by the ICPG, please refer to Appendix~\ref{prompt-design} (prompts B and C).

\subsection{Few-Shot Sampler}
We introduce a module designed to extract training examples and incorporate them into the ICL prompt. This module has access to a Few-Shot Pool containing available examples and employs techniques to select meaningful few-shot samples, providing valuable information to the LLM. This sampler can just be random, but in practice, we employ a Semantic Few-Shot Sampling technique, which finds samples based on embedding similarity with the test batch.

A Few-Shot Pool is used to retrieve few-shot examples during inference. It comprises a subset of the training dataset, containing $10\%$ of the samples for each known intent, for all known intents $n$. We introduce Semantic Few-Shot Sampling (SFS) to select relevant samples based on text similarity. We draw from Retrieval Augmented Generation (RAG) literature~\cite{lewis2020retrieval, izacard2022few} to perform a K-Nearest Neighbors (KNN) semantic sampling. We use SentenceBert~\citep{reimers2019sentence} (SBERT) to extract text embeddings and employ cosine distance to determine those examples close in the latent space. The selected examples are injected into the prompt as in-context examples.

\subsection{Known Intent Feedback}
During inference with the test set, the system will discover novel intents. Accessing both known and discovered intents is crucial for making accurate predictions. We propose using \textit{Known Intent Feedback} (KIF), a straightforward technique that conditions the LLM to reuse intents from the database by injecting the known and discovered intents directly into the prompt. This allows a frozen LLM to learn about new user queries during inference. 

Being $n$ the number of known intents, and $m$ the number of discovered intents at the end of an iteration, the system updates the database of intents to a size $n+m$. This operation makes the model aware of the accessible intents (known and discovered) and allows it to reuse them. With this operation, we are also restricting the action space of the LLM, hence regularizing the method for variations of an intent, or growing the list without control. As the LLM is promoted to discover new intents, the list of intents will grow. This process is shown in Figure~\ref{fig:figure1} and ~\ref{fig:figure2}. See Appendix~\ref{prompt-design} for the specific prompt used.

\vspace{3px}
\textbf{Semantic Known Intent Feedback Sampling (SKIF).} Following the same technique described in SFS, we incorporate the option to reduce the list of known intents included in the prompt using KNN. This introduces a variable, denoted as $n_{\text{SKIF}}$, which determines the number of intents injected into the prompt to inform the model about the current known and discovered intents. This process serves two main purposes. First, it aims to introduce ICL examples that are semantically similar to the current test batch. Second, it optimizes the use of the language model's context length by avoiding the injection of the entire list of known intents into the prompt. This approach allows us to strike a balance between metric performance and query efficiency. If this feature is not activated, all known intents pass through directly to the prompt.

\subsection{Intent Predictor}
The Intent Predictor LLM is conditioned with the generated In-Context Prompt encompassing a task description, few-shot examples, and the unique set of known and discovered intents. Test examples are introduced at the end of the prompt to facilitate inference. This prompt is constructed using a template that precisely defines its content and format (see prompt at Appendix~\ref{prompt-design}). To ensure accurate predictions, we include explicit instructions in the prompt, specifying that the model should solely focus on predicting test labels and avoid extracting few-shot samples. We also define the desired output format to simplify parsing. After obtaining predictions, we check for newly discovered intents and incorporate them into the \textit{Known Intents} database, expanding the list of intents considered by the model moving forward.

\vspace{3px}
\textbf{Clustering of Intents.} We adopt the standard choice~\citep{zhang2021discovering} of using K-Means for clustering intents. We compute SBERT embeddings on predicted and ground truth intents and perform K-Means clustering. We automatically compute the $K$ in K-Means by first using DBSCAN~\citep{ester1996density} with $\epsilon=0.5$ on the predicted embeddings.
\begin{table*}[ht]
    \centering
    \resizebox{1.0\textwidth}{!}{%
    \begin{tabular}{@{}llcccccc@{}} 
        \toprule
        && \multicolumn{3}{c}{\textbf{CLINC}} & \multicolumn{3}{c}{\textbf{BANKING}} \\
        \cmidrule(lr){3-5} \cmidrule(lr){6-8}
        \textbf{Setting} & \textbf{Method} & \textbf{NMI} $\uparrow$ & \textbf{ARI}$\uparrow$ & \textbf{ACC}$\uparrow$ & \textbf{NMI} $\uparrow$ & \textbf{ARI}$\uparrow$ & \textbf{ACC}$\uparrow$ \\
        \midrule
        \multirow{10}{*}{Unsupervised} 
                                      & KM~\citep{macqueen1967some} & 70.89 & 26.86 & 45.06 & 54.57 & 12.18 & 29.55 \\
                                      & AG~\citep{gowda1978agglomerative} & 73.07 & 27.70 & 44.03 & 57.07 & 13.31 & 31.58 \\
                                      & SAE-KM~\citet{xie2016unsupervised} & 73.13 & 29.95 & 46.75 & 63.79 & 22.85 & 38.92 \\
                                      & DEC~\citet{xie2016unsupervised} & 74.83 & 27.46 & 46.89 & 67.78 & 27.21 & 41.29 \\
                                      & DCN~\citep{yang2017towards} & 75.66 & 31.15 & 49.29 & 67.54 & 26.81 & 41.99 \\
                                      & DAC~\citep{chang2017deep} & 78.40 & 40.49 & 55.94 & 47.35 & 14.24 & 27.41 \\
                                      & DeepCluster~\citep{caron2018deep} & 65.58 & 19.11 & 35.70 & 41.77 & 8.95 & 20.69 \\
                                      & IDAS~\citep{de2023idas} & - & - & - & 80.43 & 53.31 & 63.77 \\ 
                                      \noalign{\vskip 1pt}
                                      \cdashline{2-8}
                                      \noalign{\vskip 3pt}
                                      & IntentGPT-Llama-2$_\text{0 shot}$ (ours) & 85.63 & 36.64 & 58.04 & 76.32 & 35.95 & 44.82\\
                                      & IntentGPT-3.5$_\text{0 shot}$ (ours) & 90.04 & 65.62 & 73.68 &  76.58 & 40.30 & 54.82\\ 
                                      & IntentGPT-4$_\text{0 shot}$ (ours) & 94.35 & 78.69 & 83.20 & 81.42 & 55.18 & 64.22 \\ 
                                      
        \midrule
        \multirow{9}{*}{\centering Semi-Supervised}   
                                      & PCK-means~\citep{basu2004active} & 68.70 & 35.40 & 54.61 & 48.22 & 16.24 & 32.66 \\
                                      & BERT-KCL~\citep{hsu2017learning} & 86.82 & 58.79 & 68.86 & 75.21 & 46.72 & 60.15 \\
                                      & BERT-MCL~\citep{hsu2019multi} & 87.72 & 59.92 & 69.66 & 75.68 & 47.43 & 61.14 \\
                                      & CDAC+~\citep{lin2020discovering} & 86.65 & 54.33 & 69.89 & 72.25 & 40.97 & 53.83 \\
                                      & BERT-DTC~\citep{han2019learning} & 90.54 & 65.02 & 74.15 & 76.55 & 44.70 & 56.51 \\
                                      & DeepAligned~\citep{zhang2021discovering} & 93.89 & 79.75 & 86.49 & 79.56 & 53.64 & 64.90 \\
                                      & DSSCC~\citep{kumar2022intent} & 87.91 & 81.09 & 87.91 & 81.24 & 58.09 & 69.82 \\ 
                                      & SCL~\citep{shen2021semi} & 94.75 & 81.64 & 86.91 & \underline{85.04} & \underline{65.43} & 76.55\\ 
                                      & IDAS~\citep{de2023idas} & 93.82 & 79.02 & 85.48 & - & - & - \\
                                      & LatentEM~\citep{zhou2023probabilistic} & 95.01 & \underline{83.00} & \textbf{88.99} & 84.02 & 62.92 & 74.03\\

        \midrule
        \multirow{3}{*}{\centering Few-Shot ICL}     
                                      & IntentGPT-Llama-2$_\text{10 shot}$ (ours) & 87.86 & 60.35 & 71.75 & 77.39 & 45.52 & 55.97\\
                                      & IntentGPT-Llama-2$_\text{50 shot}$ (ours) & 91.71 & 68.10 & 77.95 & 81.56 & 56.68 & 69.83\\
                                      & IntentGPT-3.5$_\text{10 shot}$ (ours) & 91.72 & 71.69 & 77.56 & 78.21 & 48.20 & 63.65\\

                                      & IntentGPT-3.5$_\text{50 shot}$ (ours) & \underline{95.60} & 76.97 & 84.86 & 81.79 & 56.60 & 68.77 \\
                                      
                                      & IntentGPT-4$_\text{10 shot}$ (ours) & 94.85 & 80.36 & 85.07 & 83.11 & 59.07 & 70.42\\
                                      & IntentGPT-4$_\text{50 shot}$ (ours) & \textbf{96.06} & \textbf{84.76} & \underline{88.76} & \textbf{85.94} & \textbf{66.66} & \textbf{77.21}\\         
        \bottomrule
    \end{tabular}
    }
    \caption{Comparison against unsupervised and semi-supervised models. The best models are shown in \textbf{bold}, and the second best are \underline{underlined}. Baseline results are extracted from~\citet{kumar2022intent}. }
    \vspace{-3mm}
    \label{tab:75KIR}
\end{table*}
 
\begin{table*}[ht]
    \centering
    \resizebox{1.0\textwidth}{!}{%
    \begin{tabular}{lccccccccccccc}
        \toprule
        &\multicolumn{5}{c}{\textbf{Prompt Features}} &\multicolumn{4}{c}{\textbf{CLINC}} & \multicolumn{4}{c}{\textbf{BANKING}} \\
        
        \cmidrule(lr){2-6} \cmidrule(lr){7-10} \cmidrule(lr){11-14}
        
        \textbf{Method} &  \textbf{KIF} & \textbf{FS} & \textbf{SFS}  & \textbf{ICP} & \textbf{SKIF} & \textbf{NMI} $\uparrow$ & \textbf{ARI} $\uparrow$ & \textbf{ACC} $\uparrow$ & \textbf{NDI} $\downarrow$ & \textbf{NMI} $\uparrow$ & \textbf{ARI} $\uparrow$ & \textbf{ACC} $\uparrow$ & \textbf{NDI} $\downarrow$ \\
        \midrule
        \multirow{7}{*}{\centering IntentGPT-3.5}  
        &  &  &  &  &  & 80.12 & 41.08 & 56.13 & 1484 & 68.26 & 21.70 & 31.74 & 997 \\
        
        & \ding{52} &  &  &  &  & 90.26 & 65.53 & 74.42 & 241 &76.58 & 40.30 & 54.82 & 120\\

        & \ding{52} & \ding{52} &  &  &  & 90.42 & \textbf{66.53} & 75.30 &  176 & 77.17 & 45.59 & 59.32 & \underline{115}\\ 

        & \ding{52} & \ding{52} & \ding{52} &  &  & 90.71 & 65.29 & 76.60 & 159 &78.79 & \underline{48.65} & \textbf{64.89} & \textbf{83}\\ 
        
        & \ding{52} & \ding{52} &  & \ding{52} &  & \underline{92.31} & 65.26 & \underline{78.04} & \textbf{149} &77.90 & 45.02 & 58.35 & 130\\ 

        & \ding{52} & \ding{52} & \ding{52} & \ding{52} &  & \textbf{93.07} & 65.39 & \textbf{78.95} & \underline{147} & \underline{80.05} & \textbf{50.12} & \underline{63.49} & 152\\ 

        & \ding{52} & \ding{52} & \ding{52} & \ding{52} & \ding{52} & 91.77 & \underline{65.62} & 75.78 & 204 & \textbf{80.54} & 43.44 & 51.06 & 574\\ 

        \midrule
        
        \multirow{7}{*}{\centering IntentGPT-4}  
        
        &  &  &  &  &  & 86.32 & 56.96 & 66.44 & 1038 & 72.00 & 30.94 & 38.12 & 1322\\

        & \ding{52} &  &  &  &  & 93.81 & 75.65 & 82.44 & \textbf{149} & 80.66 & 53.47 & 64.25 & 102\\

        & \ding{52} & \ding{52} &  &  &  & 93.58 & 76.55 & 83.73 & 155 & 81.70 & 55.09 & 64.51 & 125\\

        & \ding{52} & \ding{52} & \ding{52} &  &  & 93.89 & 75.72 & 82.09 & \underline{146} & \underline{81.71} & \underline{55.91} & \underline{67.92} & \textbf{79}\\ 
        
        & \ding{52} & \ding{52} &  & \ding{52} &  & \underline{93.95} & \underline{77.71} & \underline{83.91} & 163 & 80.54 & 54.80 & 66.40 & \underline{110}\\ 

        & \ding{52} & \ding{52} & \ding{52} & \ding{52} &  & \textbf{94.99} & \textbf{80.48} & \textbf{85.42} & 159 & \textbf{83.18} & \textbf{60.10} & \textbf{70.42} & 118\\ 

        & \ding{52} & \ding{52} & \ding{52} & \ding{52} & \ding{52} & 91.70 & 73.20 & 78.80 & 237 & 80.62 & 54.26 & 64.90 & 173\\ 
        \bottomrule
    \end{tabular}%
    }

    \caption{Ablation study of the proposed prompt features. \textbf{KIF}: Known Intent Feedback, \textbf{SKIF}: Semantic Known Intent Feedback, \textbf{SFS}: Semantic Few-Shot, \textbf{ICP}: In-Context Prompt. For NDI, the best results are the closest to the ground truth (150 for CLINC and 77 for BANKING).}
    \vspace{-3mm}
    \label{tab:ablation}
\end{table*}

\section{Experimental Setting}\label{experimental-setting}
In this section, we describe the datasets, the evaluation framework, and the Intent Discovery metrics.

\vspace{3px}
\textbf{Datasets.} We evaluate IntentGPT on several datasets. CLINC~\citep{larson2019evaluation} contains 150 intents across 10 different domains, including work, travel, and banking. It contains 18,000 train, 2,250 validation, and 2,250 test examples. BANKING~\citep{casanueva2020efficient} focuses exclusively on the finance domain, encompassing 77 distinct intents. It contains 9,003 train, 1,000 validation, and 3,080 test examples. We also evaluate our method on datasets with less number of intents like SNIPS~\cite{coucke2018snips} and StackOverflow~\cite{xu2015short}, and multilingual data~\cite{li2020mtop}. (See results in appendix~\ref{multilingual}, \ref{additional-results})

\vspace{3px}
\textbf{Evaluation setting and metrics.} We adopt the evaluation framework proposed by~\citet{zhang2021discovering}, which defines a Known Intent Ratio (KIR) and selects a percentage of training samples per each known intent, typically set at 10\% i.e. sew-shot pool. We use $n_{\text{shots}}$ number of examples in the prompt, a much lower-data setup compared to previous works. In the context of Intent Discovery, evaluation metrics pivot towards the utilization of clustering-based measures, which assess the models' effectiveness in assigning utterances to their corresponding intents. We adopt Normalized Mutual Information (NMI), Clustering Accuracy (ACC), and Adjusted Rand Index (ARI) metrics as proposed in the literature~\citep{kumar2022intent}. These metrics quantify the degree of alignment between predictions and the ground truth following the clustering of intent representations, i.e., text embeddings~\cite{lin2020discovering}. We also define the Number of Discovered Intents (NDI) as a metric to assess the deviation of the total number of discovered intents against the ground truth.

\section{Results and Discussion}\label{results}
We present the results of IntentGPT when powered by three different pre-trained LLMs, namely GPT-3.5, GPT-4, and Llama-2. See Appendix~\ref{llms} for details on these LLMs. Here we compare our method against the baselines on the CLINC and BANKING test sets, using the popular setting of varying the KIR. We also present an ablation of our method. Find in Appendix~\ref{additional-results} results on other datasets and interesting settings. During inference, we use a batch size of 16, which offers a good trade-off between context length and inference cost.

\paragraph{Effectiveness of IntentGPT}
The main result of our paper is presented here. We evaluate the performance of IntentGPT models on the $\text{KIR}=0.75$ setting, a commonly used configuration for Intent Discovery evaluation. Results are presented in Table~\ref{tab:75KIR}. We report the performance of our method in the 0-shot, 10-shot, and 50-shot settings. For the 0-shot setting, we do not use any few-shot samples, the prompt is simply a basic description of the task, and we activate Known Intent Feedback (KIF) to reuse intents. In the 10 and 50-shot settings, we utilize Semantic Few-Shot Sampling (SFS) and KIF at each iteration. Notably, IntentGPT-4 in the 50-shot setting outperforms prior methods in both unsupervised and semi-supervised scenarios, except for ACC on CLINC where LatentEM achieves a slightly higher score. IntentGPT-3.5 shows strong results competing with semi-supervised methods, both using 10 and 50-shot. Even in the 0-shot setting, which is considered unsupervised, our models outperform all previous methods and demonstrate overall competitive performance. One could argue that IntentGPT is so effective due to the underlying model being possibly trained in the datasets, we discuss this matter in Appendix~\ref{gpt-trained-on-data}.

\begin{figure*}[h]
    \centering
    \includegraphics[width=1.0\textwidth]{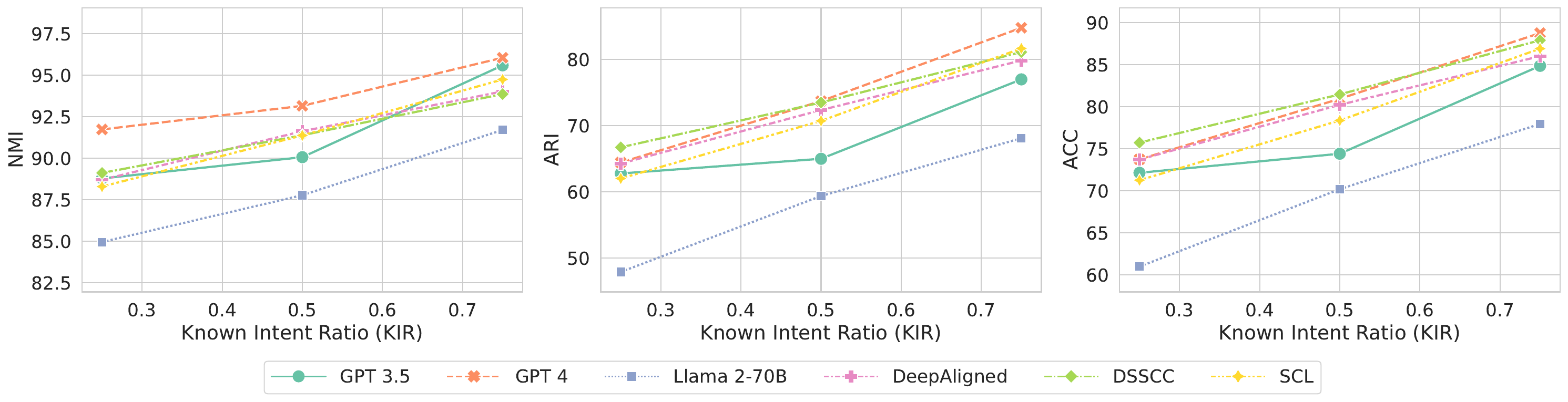} 
    \vspace{-4mm}
    \caption{Results on CLINC of IntentGPT models compared with previous baselines, as a function of the Known Intent Ratio (KIR). Ours are GPT-3.5, GPT-4 and Llama-2 70B.}
    \label{fig:curve-kir-clinc}
\end{figure*}

\begin{figure*}[h]
    \centering
    \includegraphics[width=1.0\textwidth, trim={0 5px 0 6px}, clip]{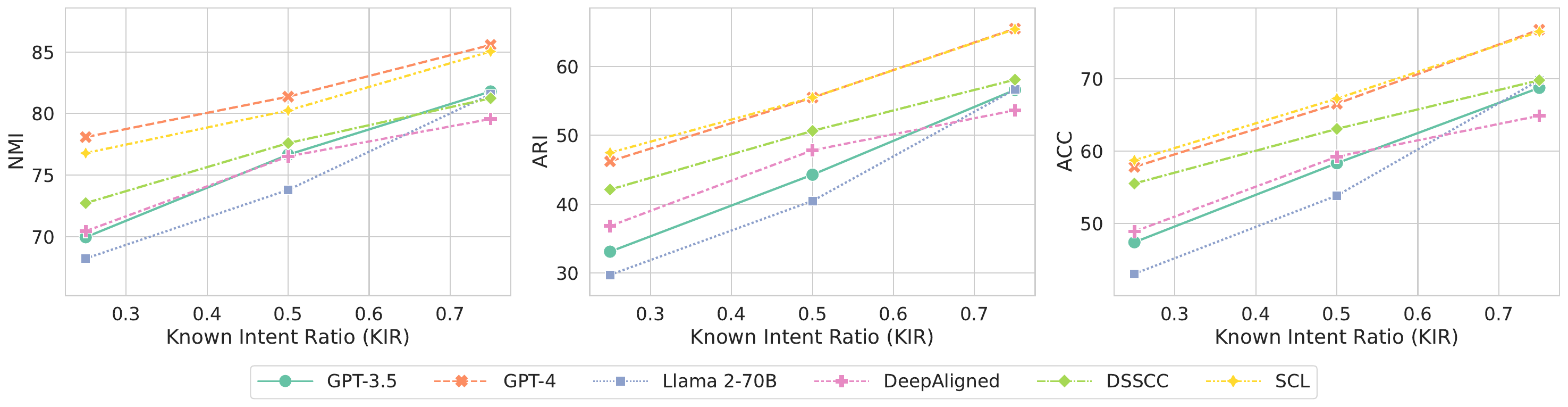}
    \vspace{-4mm}
    \caption{Results on BANKING of IntentGPT models compared with previous baselines, as a function of the Known Intent Ratio (KIR). Ours are GPT-3.5, GPT-4 and Llama-2 70B}
    \label{fig:curve-kir-banking}
\end{figure*}

\paragraph{Impact of Known Intent Ratio (KIR)}
Here we show how IntentGPT performance changes with variations in the initial number of known intents (KIR). We compare few-shot and semi-supervised methods, on three settings $\text{KIR} \in \{0.25, 0.50, 0.75\}$. Results are presented in Figures~\ref{fig:curve-kir-clinc} and~\ref{fig:curve-kir-banking}. IntentGPT outperforms all previous methods in all KIR settings. We can observe a clear performance scale when comparing LLMs. GPT-4 is the winner, followed by GPT-3.5 and Llama-2. This is consistent with results seen in popular NLP benchmarks\footnote{\url{https://llm-leaderboard.streamlit.app/}}. Due to the scale in network parameters, and training procedure, GPT-4 (with our prompting scheme) excels on Intent Discovery.

\paragraph{Ablation on Prompt Features}\label{ablation}
We investigate the influence of the proposed prompt features on the performance of IntentGPT using GPT-3.5 and GPT-4 models.  We use the $\text{KIR}=0.75$ setting. When activating Few-Shot (FS) we introduce 10 shots, and when applying Semantic Known Intent Feedback (SKIF), we inject the 50 known intents that exhibit the highest similarity to the test utterances. Table~\ref{tab:ablation} presents the results for the most significant combinations of prompt features, providing insights into their influence. Notably, the inclusion of Known Intent Feedback (KIF) yields substantial gains, as its absence results in a significant increase in the number of discovered intents, hence failing in other metrics. This is attributed to the model not being aware of the known intents and its inability to reuse them. The utilization of few-shot learning (FS) demonstrates its advantages, particularly when applying Semantic Few-Shot Sampling (SFS). It is worth noting that while SKIF does not result in significant gains, it aligns with expectations, as the optimal scenario involves including all known intents. Notably, we see performance improvements when using the automatic prompt generation (ICP) module.


\paragraph{Limitations} \label{limitations}
We outline the limitations of our model. We have a strong restriction on the context length of the LLM, and the context limit may vary over models. The attention operation makes computational and monetary costs increase as context increases. Much research is being held in this matter~\cite{bulatov2023scaling, dao2022flashattention}, hence future models might not suffer from this issue. Running the model on large LLMs like GPT-4 can be expensive and raises concerns about data management through the API, potentially impacting privacy and security. In contrast, Llama's ability to run on local machines provides a cost-effective and more secure alternative.

\section{Conclusion}\label{conclusion}
In this paper, we explore the problem of Intent Discovery by harnessing the In-Context learning capabilities of Large Language Models like GPT-3.5, GPT-4, and Llama-2. We propose IntentGPT as a general framework for discovering novel user intents, that automatically generates context-aware prompts using LLMs coupled with advanced prompting techniques like Semantic Few-Shot Sampling or Known Intent Feedback. We highlight IntentGPT's outperformance over other unsupervised and semi-supervised methods, which is particularly remarkable considering that previous approaches require larger training data and multi-stage training processes. In contrast, we do not need training and only a few demonstrations. We present exhaustive experimentation in the novel few-shot In-Context Learning setting, and ablations to understand the influence of hyperparameters. We aim to spark a new avenue for other open-set classification scenarios, such as those encountered in computer vision, by harnessing the power of Visual-Language Models.

\paragraph{Ethics Statement}\label{ethics}
We acknowledge that biases in user utterances may persist or amplify our model's inferred intents, as we do not explicitly eliminate these biases. We depend on red-teaming efforts to identify and address bias in LLMs. While biases and harmful content filtering are well-addressed on GPT models, concerns persist with LLMs like Llama due to differences in data, training, and deployment techniques, potentially leading to unfair content generation. Mitigating these concerns is a crucial aspect of responsible AI research and development.

\paragraph{Use of AI Assistants}
AI assistants were used in this work for improving productivity of code development and aid in the manuscript writing.

\bibliography{acl_latex}

\newpage

\appendix

\section{Baselines for Intent Discovery}

Similar to~\citet{kumar2022intent}, we compare our method to unsupervised approaches, including K-Means (KM)~\citep{macqueen1967some}, Agglomerative Clustering (AG)~\citep{gowda1978agglomerative}, DEC and SAE-KM by~\citet{xie2016unsupervised}, DCN~\citep{yang2017towards}, DAC~\citep{chang2017deep} and, DeepCluster~\citep{caron2018deep}. For the semi-supervised setting, we use PCK-means~\citep{basu2004active}, BERT-KCL~\citep{hsu2017learning}, BERT-MCL~\citep{hsu2019multi}, CDAC+~\citep{lin2020discovering}, BERT-DTC~\citep{han2019learning}, DeepAligned~\citep{zhang2021discovering}, DSSCC~\citep{kumar2022intent}, SCL~\citep{shen2021semi}, and LatentEM~\citep{zhou2023probabilistic}. IDAS~\citep{de2023idas} is also evaluated in a hybrid fashion. Additionally, we evaluate IntentGPT in a few-shot ICL setting, which uses significantly fewer training examples compared to the semi-supervised setting and is more comparable to an unsupervised scenario.

\section{Additional Results and Datasets} \label{additional-results}
To validate the robustness of IntentGPT we present results on additional datasets. We select SNIPS and StackOverflow as challenging datasets that contain a small number of intents. Further, we test the multilingual capabilities of IntentGPT by using the MTOP multilingual dataset. We also share an ablation on the number of shots used for few-shot ICL in different KIR regimes.

\subsection{SNIPS and StackOverflow}
Here we present results on SNIPS~\cite{coucke2018snips} and StackOverflow~\cite{xu2015short}. These are datasets that contain a much-reduced number of intents compared to CLINC or BANKING, therefore we can judge the robustness of IntentGPT in this regime. SNIPS contains over 16,000 data samples of user queries ranging 7 different intents. StackOverflow dataset contains over 3,3 million examples with 20 different intents. We do not fine-tune any models on this data. We select a few examples following the described procedure to perform Intent Discovery. We make use of 10-shot and KIR=0.75. Results are shown in Table~\ref{tab:SNIPS-StackOverflow}. IntentGPT outperforms previous state-of-the art methods in all metrics except of Clustering Accuracy (ACC) for StackOverflow, where DSSCC shows slightly better scores. Notably, it uses orders of magnitude less training data and no training. This result showcases the strength of our method into being able to process multiple datasets with no modifications on the further intervention, as it will automatically generate the prompts customized on the given data.

\subsection{Multilingual datasets}\label{multilingual}
We evaluate IntentGPT's robustness when being used for languages other than English. To that end, we use MTOP~\cite{li2020mtop} dataset, composed of utterances and intents in six different languages: English, German, French, Spanish, Hindi, and Thai. We evaluate IntentGPT in a 10-shot setting and using KIR=0.75. Results are shown in table \ref{tab:MTOP}

\begin{table}[t]
    \centering
    \begin{tabular}{@{}lccc@{}} 
        \toprule
        \textbf{Model} & \textbf{NMI} & \textbf{ARI} & \textbf{ACC} \\
        \midrule
        \multicolumn{4}{c}{\textbf{MTOP English}}  \\
        \midrule
        IntentGPT-3.5 & 83.54 & 80.43 & 76.47  \\
        IntentGPT-4 & 87.98 & 82.22 & 85.09\\
        \midrule
        \multicolumn{4}{c}{\textbf{MTOP Spanish}}  \\
        \midrule
        IntentGPT-3.5 & 88.21 & 88.74 & 82.42  \\
        IntentGPT-4 & 90.37 & 90.90 & 85.69\\
        \midrule
        \multicolumn{4}{c}{\textbf{MTOP French}}  \\
        \midrule
        IntentGPT-3.5 & 88.97 & 88.44 & 83.43  \\
        IntentGPT-4 & 84.34 & 80.62 & 76.69\\
        \midrule
        \multicolumn{4}{c}{\textbf{MTOP German}}  \\
        \midrule
        IntentGPT-3.5 & 80.44 & 65.82 & 67.26  \\
        IntentGPT-4 & 86.64 & 81.87 & 77.54\\
        \midrule
        \multicolumn{4}{c}{\textbf{MTOP Hindi}}  \\
        \midrule
        IntentGPT-3.5 & 60.43 & 36.20 & 43.33  \\
        IntentGPT-4 & 73.23 & 66.36 & 66.01\\
        \midrule
        \multicolumn{4}{c}{\textbf{MTOP Thai}}  \\
        \midrule
        IntentGPT-3.5 & 28.38 & 5.66 & 20.66  \\
        IntentGPT-4 & 23.72 & 7.42 & 17.67\\
        \bottomrule
    \end{tabular}
    \caption{Results on MTOP dataset with 6 different languages. IntentGPT with GPT-4 shows good performance at English, Spanish, French, and German. Language with less presence in the training dataset of GPT-4 are penalized, like Hindi or Thai.}
    \label{tab:MTOP}
\end{table}
\begin{table*}[t]
    \centering
    \begin{tabular}{@{}lccccccc@{}} 
        \toprule
        & & \multicolumn{3}{c}{\textbf{SNIPS}} & \multicolumn{3}{c}{\textbf{StackOverflow}} \\
        \cmidrule(lr){3-5} \cmidrule(lr){6-8}

        \textbf{Method} & \textbf{KIR} & \textbf{NMI} $\uparrow$ & \textbf{ARI}$\uparrow$ & \textbf{ACC}$\uparrow$& \textbf{NMI} $\uparrow$ & \textbf{ARI}$\uparrow$ & \textbf{ACC}$\uparrow$ \\
        \midrule

           DSSCC & 0.75 & 90.44 & 89.03 & 94.87 & 77.08 & 68.67 & \textbf{82.65} \\

            CDAC+ & 0.75 & 89.30 & 86.82 & 93.63 & 69.46 & 52.59 & 73.48\\
       \midrule                           
            IntentGPT-3.5 (ours) & 0.75 & 85.26 & 80.48 & 89.14 & 80.56 & 71.57 & 81.69\\
                                           
            IntentGPT-4 (ours) & 0.75 & \textbf{91.42} & \textbf{90.63} & \textbf{94.89} & \textbf{81.78} & \textbf{76.85} & 81.75 \\
                                            
        \bottomrule
    \end{tabular}
    
\caption{Results on SNIPS and StackOverflow datasets. IntentGPT is improves upon baselines across metrics, except for ACC on StackOverflow.}
\vspace{-2mm}
\label{tab:SNIPS-StackOverflow}
\end{table*}

We observe that IntentGPT with both GPT3.5 and GPT4 seamlesly generalizes to languages other than English without requiring extra human effort. Notably, the first LLM, in charge of generating the prompt (ICPG), automatically generates the prompt in the desired language by looking at few training examples (see the generated prompts in Spanish and French). The process of generating the prompt is explained in Section~\ref{method}.

Quantitative results show that our method works well in this multilingual setting, showing metrics in a similar range to the ones seen in other benchmarks, except for the case of Thai, where the GPT3.5 and GPT4 models struggle to handle. We leave as future work the comparison of IntentGPT with existing methods in this benchmark since existing methods do not report results on these datasets. These results highlight the robustness of IntentGPT open-world class discovery tasks, and the importance of the ICPG for automatically generating informative prompts. 

\subsection{Impact of increasing the number of few-shots}
We conduct a set of experiments where we increase the number of few-shot examples injected in the prompt for ICL, to asses the gains we can obtain when showing more context examples to the LLM. We perform these experiments on all the KIR settings and compare them against two state-of-the-art baselines, namely DeepAligned and SCL which report results on the same settings. Results are depicted in Table~\ref{tab:increase-shots}. Our observations indicate that increasing the number of shots leads to significant performance improvements, particularly in the KIR=0.75 setting. However, escalating the number of shots in the KIR = 0.25 scenario does not yield proportionate performance growth. This phenomenon is attributed to the model's limited number of absolute intents, leading to divergence over time.

\begin{table*}[ht]
    \centering
    \resizebox{1.0\textwidth}{!}{%
    \begin{tabular}{@{}llccccccccc@{}} 
        \toprule
        &&& \multicolumn{4}{c}{\textbf{CLINC}} & \multicolumn{4}{c}{\textbf{BANKING}} \\
        \cmidrule(lr){4-7} \cmidrule(lr){8-11}

        \textbf{KIR} & \textbf{Method} & \textbf{Shots} & \textbf{NMI} $\uparrow$ & \textbf{ARI}$\uparrow$ & \textbf{ACC}$\uparrow$ & \textbf{NDI} $\downarrow$ & \textbf{NMI} $\uparrow$ & \textbf{ARI}$\uparrow$ & \textbf{ACC}$\uparrow$ & \textbf{NDI} $\downarrow$\\
        \midrule

        \multirow{7}{*}{\centering 0.25} 
                                         & DeepAligned & - & 88.71 & 64.27 & \underline{73.71} & - & 68.88 & 35.49 & 47.58 & -\\
                                         & SCL & - & 88.30 & 62.02 & 71.25 & - & 76.79 & \textbf{47.47} & \underline{58.73} & -\\
        
                                         \noalign{\vskip 1pt}
                                         \cdashline{2-11}
                                         \noalign{\vskip 3pt}
        
                                        &\multirow{5}{*}{IntentGPT-4 (ours)}
                                        & 10 & 88.03 & 55.29 & 66.93 & \underline{136} & 76.25 & 42.39 & 55.91 & 99\\
                                        & & 20 & \underline{91.13} & \textbf{66.57} & \textbf{74.18} & 169 & 76.81 & 43.95 & 55.58 & 90\\
                                        & & 30 & \textbf{91.74} & \underline{64.43} & \underline{73.73} & \underline{136} & \underline{76.84} & 44.30 & 56.43  & 94\\
                                        & & 40 & 90.45 & 59.59 & 69.11  & \textbf{142} & 76.09 & 40.14 & 52.60  & \textbf{75} \\
                                        & & 50 & 90.09 & 58.09 & 70.31 & \underline{136} & \textbf{78.73} & \underline{46.54} & \textbf{59.09} & \underline{89}\\
        
        \midrule
        
        \multirow{7}{*}{\centering 0.5}  
                                        & DeepAligned & - & 91.63 & 72.34 & 80.22 & - & 76.14 & 47.07 & 59.44  & - \\
                                        & SCL & - & 91.38 & 70.71 & \underline{78.36} & - & 80.25 & \textbf{55.50} & \textbf{67.28}  & -\\
                         
                                        \noalign{\vskip 1pt}
                                        \cdashline{2-11}
                                        \noalign{\vskip 3pt}
                                        
                                        & \multirow{5}{*}{IntentGPT-4 (ours)} 
                                        & 10 &  91.44 & 66.93 & 74.44 & \underline{143} & \textbf{81.06} & \underline{55.17} & 66.33 & 131\\
                                        & & 20 & 91.41 & 63.16 & 73.11 & 141 & 79.77 & 52.50 & 64.97  & 111\\
                                        & & 30 & 92.72 & \underline{72.47} & 79.16 & 141 & \underline{80.43} & 53.21 & \underline{66.79} & 102\\
                                        & & 40 & \underline{92.93} & 71.96 & 77.47  & 140 & 79.71 & 51.26 & 63.02  & \textbf{88}\\
                                        & & 50 & \textbf{93.16} & \textbf{73.69} & \textbf{80.94}  & \textbf{155} & 79.45 & 50.22 & 62.76  & \underline{92}\\
        \midrule
        
        \multirow{7}{*}{\centering 0.75}  
                                        & DeepAligned & - & 94.03 & 79.82 & 86.01 & - & 78.77 & 52.11 & 63.68   & -\\
                                        & SCL & - & 94.75 & 81.64 & 86.91  & - & \underline{85.04} & \underline{65.43} & \underline{76.55}  & -\\
        
                                        \noalign{\vskip 1pt}
                                        \cdashline{2-11}
                                        \noalign{\vskip 3pt}
                                        
                                        & \multirow{5}{*}{IntentGPT-4 (ours)}
                                        & 10 & 94.69 & 79.56 & 85.07 & 164 & 83.46 & 59.91 & 70.39 & 109\\
                                        & & 20 & 95.22 & 81.51 & 86.22 & 163 & 84.49 & 63.73 & 74.03 & 114\\
                                        & & 30 & 95.75 & 83.89 & 87.82 & 166 & \textbf{85.32} & 65.30 & 76.33 & 98\\
                                        & & 40 & \underline{95.97} & \underline{84.08} & \underline{88.49} & \textbf{156} & 85.01 & 64.07 & 74.58 & \underline{95}\\
                                        & & 50 & \textbf{96.06} & \textbf{84.76} & \textbf{88.76} & \underline{159} & \underline{85.07} & \textbf{66.66} & \textbf{77.21} & \textbf{88}\\
        \bottomrule
    \end{tabular}
    
    }
\caption{Performance gains of IntentGPT when we increase the number of shots for few-shot ICL using GPT-4. Baseline results are extracted from~\citet{kumar2022intent} and ~\citet{zhang2021discovering}.}
\vspace{-2mm}
\label{tab:increase-shots}
\end{table*}

\subsection{Computation of Intent Embeddings for Clustering}
We compute embeddings of predicted intents to perform clustering. Here we show an ablation study to decide the best type of embeddings for our task. We experiment with OpenAI embeddings \texttt{text-embedding-ada-002} \footnote{\url{https://platform.openai.com/docs/guides/embeddings}}. Results are shown in Table~\ref{tab:embeddings}. We find that SentenceBert outperforms OpenAI embeddings. The most likely reason for OpenAI embeddings being worse is that it is primarily used to encode large contexts like documents for document retrieval, whereas SentenceBert is trained to efficiently encode short contexts, or sentences, like the size of our intents.

\begin{table*}[t]
    \centering
    \begin{tabular}{@{}llcccccccc@{}} 
        \toprule
        && \multicolumn{3}{c}{\textbf{CLINC}} & \multicolumn{3}{c}{\textbf{BANKING}} \\
        \cmidrule(lr){3-5} \cmidrule(lr){6-9}

        \textbf{Method} & \textbf{Embedding type} & \textbf{NMI} $\uparrow$ & \textbf{ARI}$\uparrow$ & \textbf{ACC}$\uparrow$& \textbf{NMI} $\uparrow$ & \textbf{ARI}$\uparrow$ & \textbf{ACC}$\uparrow$ \\
        \midrule

           IntentGPT-3.5 & SentenceBert & 91.66 & 71.58 & 78.99 & 88.21 & 88.74 & 82.42 \\

            IntentGPT-3.5 & text-embedding-ada-002 & 88.64 & 61.42 & 71.36 & 69.46 & 39.51 & 52.12\\
                                          
            IntentGPT-4 & SentenceBert & 94.01 & 77.17 & 83.96 & 91.66 & 71.58 & 78.99\\
                                           
            IntentGPT-4 & text-embedding-ada-002 & 90.73 & 65.99 & 74.89 & 88.64 & 61.42 & 71.36 \\
                                            
        \bottomrule
    \end{tabular}

    \vspace{-3mm}
\caption{Results of using different types of embeddings for clustering intents}
\vspace{-2mm}
\label{tab:embeddings}
\end{table*}

\subsection{Influence of $K$ for K-Means clustering}
\begin{figure*}[h]
    \centering
    \includegraphics[width=1.0\textwidth]{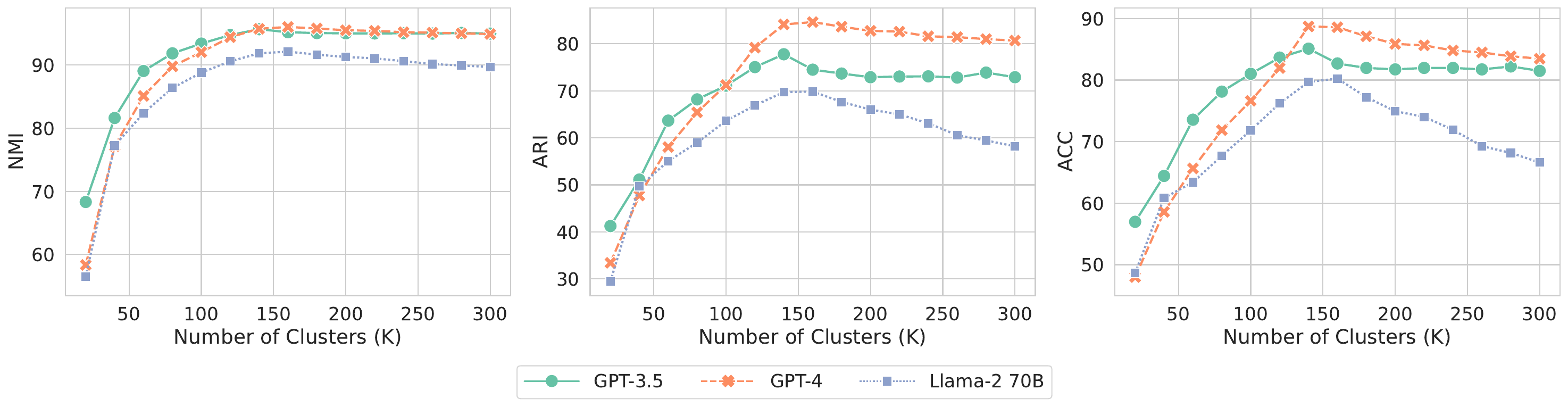}
    \vspace{-4mm}
    \caption{Analyzing the influence of $K$ for K-Means on CLINC dataset (150 intents).}
    \label{fig:K-clinc}
\end{figure*}

\begin{figure*}[h]
    \centering
    \includegraphics[width=1.0\textwidth]{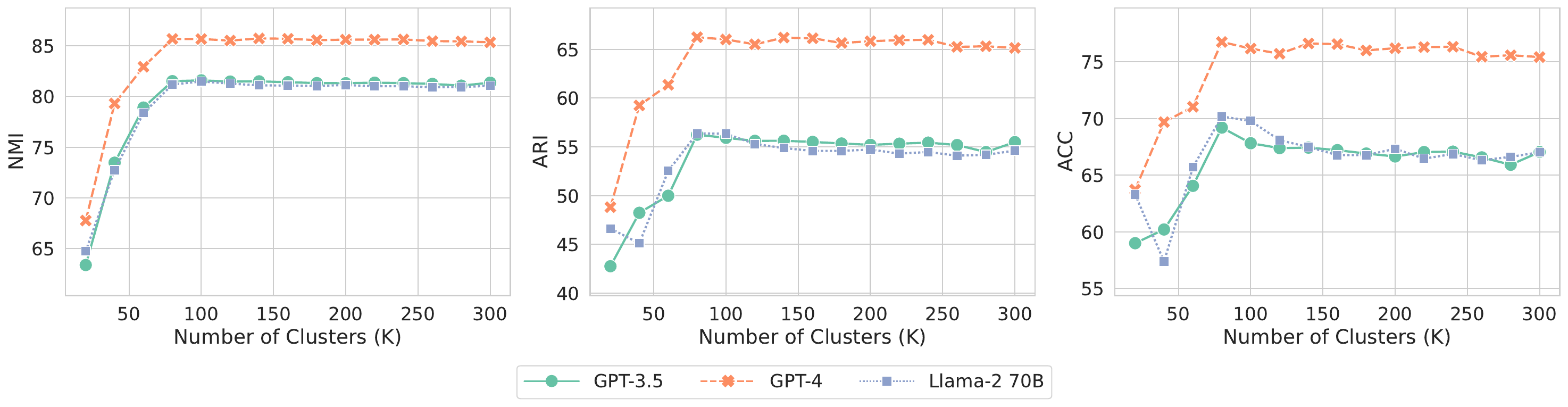}
    \vspace{-4mm}
    \caption{Analyzing the influence of $K$ for K-Means on BANKING (77 intents).}
    \label{fig:K-banking}
\end{figure*}

We explore the choice of $K$ in K-Means under the KIR=0.75 setting. We explore all IntentGPT models in the 50 shot setting. Results are shown in Figures~\ref{fig:K-clinc} and ~\ref{fig:K-banking}. We observe that there exists an optimal range where the models perform at their best, typically aligning with the ground truth number of intents in the datasets, approximately $K=150$ for CLINC and $K=77$ for BANKING. It's worth noting that GPT models tend to saturate in metric performance as $K$ increases, while Llama-2 exhibits a slight decline. 

In our IntentGPT experiments, we automatically determine the value of $K$ for K-Means by first using DBSCAN on predicted intent embeddings. This approach aligns with real-world scenarios where the total number of intents is unknown. We find that $K$ tends to be approximately 155 for CLINC and 88 for BANKING.

\subsection{Sampling Temperature of LLMs}
The sampling temperature quantifies the desired level of stochasticity during the generation process, typically represented as a positive scalar within the 0 to 2 range. Higher temperatures foster greater creativity in the model, while lower temperatures favor determinism. In our specific context, our objective is to foster creativity, enabling the model to uncover novel and precise intents while maintaining a connection with existing intents. 

\begin{figure*}[h]
    \centering
    \includegraphics[width=1.0\textwidth]{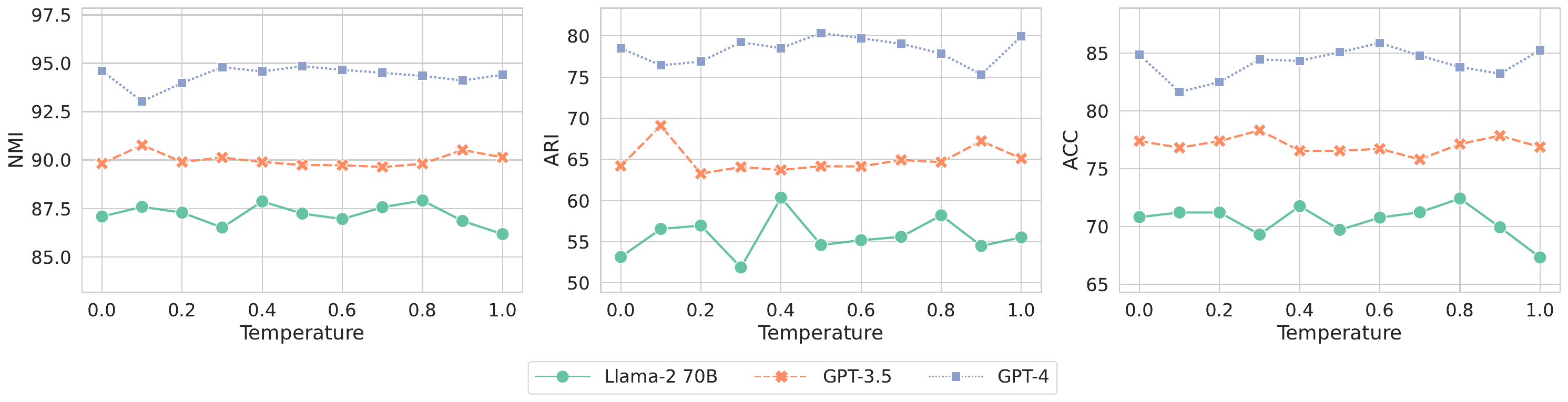}
    \vspace{-4mm}
    \caption{Analyzing the influence of the LLM temperature on CLINC dataset.}
    \label{fig:temp-clinc}
\end{figure*}

\begin{figure*}[h]
    \centering
    \includegraphics[width=1.0\textwidth]{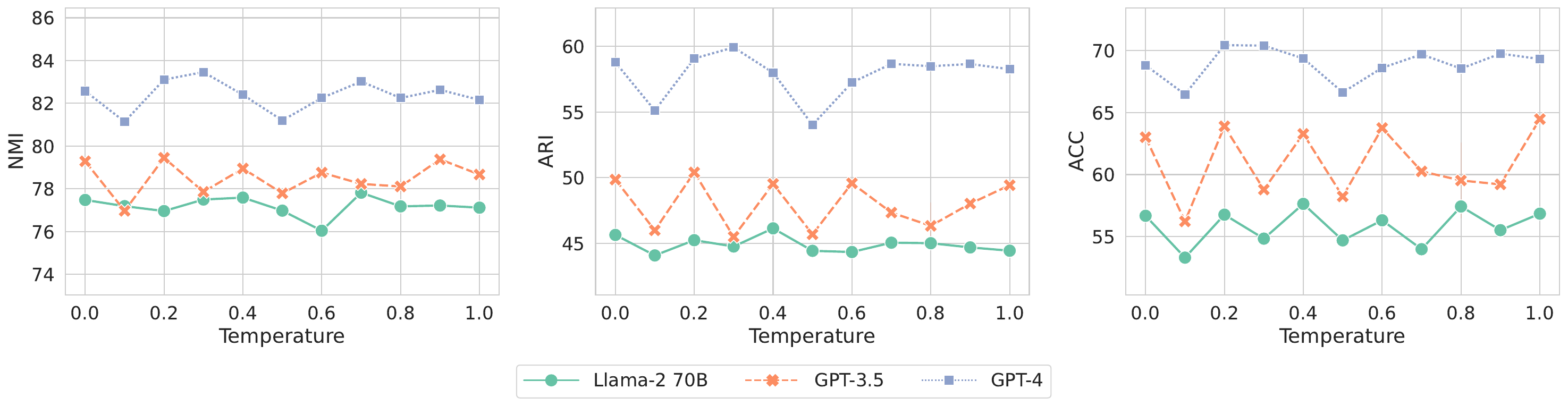}
    \vspace{-4mm}
    \caption{Analyzing the influence of the LLM temperature on BANKING dataset.}
    \label{fig:temp-banking}
\end{figure*}

In our evaluation, depicted in Figures~\ref{fig:temp-clinc} and~\ref{fig:temp-banking}, we observe a trend of declining metrics as temperature increases. This phenomenon occurs because the model's increased randomness at higher temperatures can lead to unintended variations in the generated intents, potentially affecting the accuracy of the model's responses. Thus, selecting an optimal temperature is crucial to strike the right balance between creativity and coherence in intent generation.

\section{IntentGPT Algorithm} \label{algorithm}
We define the algorithm of the main operations of IntentGPT in Algorithm~\ref{intent-gpt-algorithm}. Our proposed system performs Few-Shot Intent Discovery by selecting relevant few-shot examples from the training dataset using Semantic Few-Shot Sampling (SFS). It then determines a subset of intents to include in the prompt, considering both known intents and potential new ones, by employing the Known Intent Feedback (KIF) mechanism. Subsequently, it generates a contextually informative prompt for the Intent Discovery task. The pre-trained Large Language Model (LLM), such as GPT-4, processes this prompt and provides a response parsed to extract predicted intents. If a predicted intent is not already known, it is added to the list of known intents, contributing to the system's ability to discover new intents over time. This algorithm demonstrates the core operations of IntentGPT for Few-Shot Intent Discovery.
\begin{algorithm*}[ht]
\caption{Few Shot Intent Discovery Algorithm}
\label{alg:text_inference}
\begin{algorithmic}[1]
\Procedure{Few Shot Intent Discovery}{$text, trainSet, knownIntents, LLM$}
\State $fewShotPool \gets $ SFS($text, trainSet$) \Comment{Select few-shot examples}
\State $intentList \gets$ SKIF$(text, knownIntents)$ \Comment{Select subset of intents to include}
\State $prompt \gets$ CreatePrompt$(fewShotPool, intentList)$ 
\State $response \gets$ LLM($prompt$)
\State $predictedIntent \gets $DiscoverNewIntent$(response)$ \Comment{Parse response for predicted intents}
\If{$predictedIntent \notin knownIntents$} \Comment{Check if the predicted intent is new to add.}
    \State $knownIntents.add(predictedIntent)$
\EndIf
\EndProcedure
\end{algorithmic}
\label{intent-gpt-algorithm}
\end{algorithm*}
\subsection{Prompt Design} \label{prompt-design}
\renewcommand\lstlistingname{Prompt}

In this section, we show the prompts we used to empower the in-context learning capabilities of IntentGPT. As described in section \ref{method}, we use GPT-4 for designing context-aware prompts. Prompt A is the initial prompt that describes the task of designing prompts using several context examples. We inject $x=2$ random samples per known intent to give context to GPT-4. The generated prompt, e.g., Prompt B or C, is used as a guideline for Intent Discovery during inference on the LLM tasked at Intent Discovery. Prompt D shows a complete interaction with GPT-4 for an iteration step of Intent Discovery. Note that the prompt contains a list of the known intents, few-shot demonstrations, and a batch of test examples. Prompt E shows a simple human-generated prompt.

\section{Detail on off-the-shelf LLMs used in IntentGPT}\label{llms}
This section describes our usage of pre-trained language models. Find here details on specific models, ways to access them, and estimated costs for running them. For GPT-4 and GPT-3.5 we use the available endpoints at OpenAI\footnote{https://platform.openai.com/playground}, namely \texttt{gpt-3.5-turbo} and \texttt{gpt-4}. For Llama-2, we use the AnyScale\footnote{https://app.endpoints.anyscale.com/} endpoint that uses \texttt{meta-llama/Llama-2-70b-chat-hf}. Table~\ref{tab:model-cost} displays relevant information about the models, the tokens required for a single query with a batch of 16 samples, and their monetary costs. Finally, we use a pre-trained SentenceBert~\citep{reimers2019sentence} transformer (\texttt{all-MiniLM-L6-v2}) model aimed at extracting text embeddings for computing text similarities and clustering.

Considering a batch size of 16 samples for query, CLINC requires 141 iterations to perform a complete test evaluation, while BANKING requires 193. For instance, Table~\ref{tab:model-cost} suggests that an evaluation run using GPT-4 costs approximately 7\$ on CLINC and 9.65\$ on BANKING. In terms of timing, we measure that GPT-3.5 takes approximately 20 minutes, while GPT-4 finishes in around 2 hours. These results show that running inference on these models can be costly, but presents a reasonable alternative to fine-tuning a model on specific cases and domains.

\begin{table*}[t]
    \centering
    \begin{tabular}{lccc} 
        \toprule
        \textbf{Model} & \textbf{Endpoint API} & \textbf{\# Tokens} & \textbf{Cost/Batch (\textdollar)} \\
        \midrule
        \texttt{gpt3.5-turbo} & OpenAI & 1240 & 0.002  \\
        \texttt{gpt-4} & OpenAI & 1240 & 0.05\\
        \texttt{Llama-2-70b-chat-hf} & AnyScale & 1518 & 0.0015 \\
        \bottomrule
    \end{tabular}
    \caption{Cost per batch inference for different models.}
    \label{tab:model-cost}
\end{table*}

\section{Are the LLMs we use Pre-Trained on the Datasets?}\label{gpt-trained-on-data}

One could argue that the effectiveness of IntentGPT arises from the underlying LLM, such as GPT-4, having been trained on the data~\citep{mireshghallah2022quantifying,mattern2023membership}. Nevertheless, addressing this question proves challenging due to a lack of transparency in this matter. To explore this further, we will conduct three analyses. First, we will perform a series of interventions on GPT-4 to determine its familiarity with the CLINC and BANKING test datasets. Second, we will revisit our ablation study on GPT-4 to gain insights into the source of improvements. Finally, we introduce a new metric to assess the similarity between the intents generated by IntentGPT using GPT-4 and the ground truth.

\subsection{GPT-4 on Intent Discovery Datasets.} We perform a series of interventions and engage in conversation with the GPT-4 model, seeking to understand how much it knows from CLINC and BANKING datasets. We ask for detailed descriptions of the datasets, generation of train and test examples, and prediction of test labels. From this study, we acknowledge that GPT-4 knows both datasets, domain contexts, number of samples and splits, and understands the task. However, we did not find that any of the examples generated were part of the datasets, hence the examples were hallucinated. 

\subsection{Further discussion on ablation study.} We revisit the ablation study from Table~\ref{tab:ablation} to understand where the gains in performance come from. The most simple set-up, where we only include a human-generated task description (first row in the table), shows very poor performance in all metrics. This shows that the model by itself is not able to perform well in the task. Is when we add the proposed techniques such as Known Intent Feedback (KIF), Semantic Few-Shot (SFS), or In-Context Prompt Generation (ICP) that the models start to achieve state-of-the-art performance.

\subsection{Measuring the similarity between generated intents and real ones.} We want to measure the similarity between the predicted and the ground truth intents, to learn if GPT-4 has memorized the test labels during training with the test set. In contrast, we would like the model to generalize enough that it creates novel intents and consistently assigns them to the corresponding utterance. To that end, we design the \textit{Frechet Bert Distance} ($\text{FBD}$) metric that takes inspiration from the well-known Fréchet Inception Distance~\cite{heusel2017gans} used in computer vision for assessing the quality of image generative models. We build on top of \citet{semeniuta2018accurate} and choose to use SBERT embeddings for computing the similarity between text distributions in the latent space. 

Having two sets of text samples, $s_1$ and $s_2$, we extract SBERT embeddings and compute the Fréchet Distance between them. If both sets are equal, we will measure zero, and if they are completely different we will measure close to one. For example, having $s_1=(\text{'a'}, \text{'b'}, \text{'c'})$ and $s_2=(\text{'d'}, \text{'e'}, \text{'f'})$,  gives $\text{FBD}(s_1, s_2) = 0.96$. 

For our experiment, we select the set of discovered intents $I_d$ (excluding the known intents at the start) and the set of ground truth unknown intents to discover $I_u$, and compute SBERT embeddings. Moreover, we compute the Fréchet Distance between both sets of embeddings. Our final measurement for the results using GPT-4 on CLINC with 50-shots is $\text{FDB}(I_d, I_u) = 0.54$. In the case of BANKING we obtain $\text{FDB}(I_d, I_u) = 0.55$, which suggests that generated intents are sufficiently different in the used datasets. Additionally, we can evaluate the generated intents qualitatively using Table~\ref{tab:qualitative-examples}, which shows also the differences between the generated and real intents.

This is a relevant problem and orthogonal to many research directions in LLMs because pre-trained models that are trained on test sets, may break the confidence of future research efforts in popular benchmarks for important problems. Future work must put effort into detecting which models can be evaluated in which dataset, and even though that is the case, evaluate the level of memorization of test predictions and model generalization.

\subsection{Qualitative analysis of the Discovered Intents}
The design of IntentGPT allows us to qualitatively assess the discovered intents because the model will inherently find the need for a new intent and create a name for it. Table~\ref{tab:qualitative-examples} displays some of the examples seen during test on CLINC with GPT-4 on KIR=0.75 and 50 shot settings. By looking at the test predictions we observe that the model is consistent in assigning the utterances to the same intent, even though the name of the intent does not correspond the the ground truth. This is the correct behavior, because clustering metrics do not take into account that discovered and ground truth intents are the same. However, we can see errors in utterances with subtle nuances like the difference between "bill inquiry" and "bill payment", which was the prediction of IntentGPT for "pay bill". Also, "device\_pairing" is different than the ground truth "sync\_device", but is consistently assigned to the corresponding utterances.

\begin{figure*}[h]
    \centering
    \includegraphics[width=0.7\textwidth]{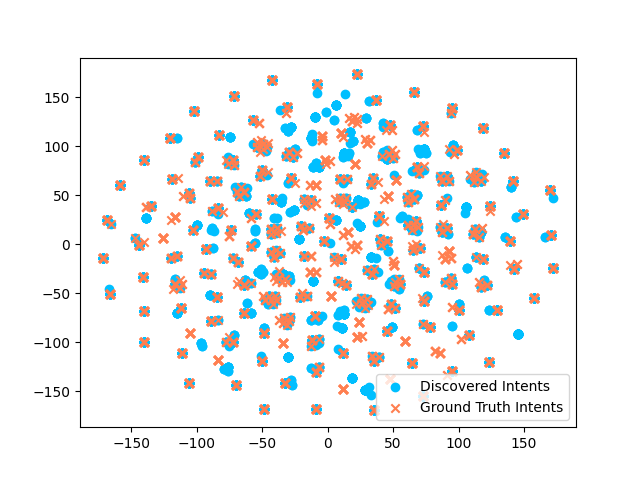} 
    \vspace{-0.5cm}
    \caption{TSNE visualization of S-Bert embeddings of discovered intents and ground truth intents. We show that GPT-4 does not memorize intents in case of hypothetical test-data leakage during its pre-training. We observe some regions with examples that match more that and others, showing a tie. That is because many intents can be naively predicted with small divergence, while others are more challenging. }
    \label{fig:tsne}
\end{figure*}

Furthermore, we compute SBert embeddings of both discovered intents and ground truth intents and compute a TSNE visualization into a 2D grid. Results are shown in Figure~\ref{fig:tsne}. This visualization displays that many intents coincide while many others are semantically different. We consider that this is consistent with the Frechet Bert Distance score obtained before and that both sets are sufficiently different.

\begin{table*}[htbp]
\centering

\resizebox{1.0\textwidth}{!}{%

\begin{tabular}{|c|c|c|}
\hline
\textbf{Utterance} & \textbf{Predicted Intent} & \textbf{Ground Intent} \\ \hline
i lost my phone and need help retrieving it & device\_location & find\>phone \\ \hline
are you able to track a phone & device\>location & find\>phone\\ \hline
call my phone, i can't find it & make\>call & find\>phone\\ \hline
at what time will my flight land & flight\>status & flight\>status\\ \hline
is flight dl123 coming in on time & flight\>status & flight\>status\\ \hline
i would like to know when my flight scheduled to board & flight\>status & flight\>status\\ \hline
choose heads or tails and tell me what you came up with & flip\>coin & flip\>coin\\ \hline
lets do a coin toss & roll\>dice & flip\>coin\\ \hline
what's the least payment i can make on my cable bill & bill\_inquiry & min\_payment\\ \hline
how much is the least i can pay for power bill & bill\_inquiry & min\_payment\\ \hline
i need to pay the smallest amount on my phone bill & bill\_inquiry & min\_payment\\ \hline
i would like to know more about getting a new credit card & credit\_card\_inquiry & new\_card\\ \hline
what's the procedure to apply for a new credit card & credit\_card\_inquiry & new\_card\\ \hline
i would like to get a new credit card & credit\_card\_inquiry & new\_card\\ \hline
does bank of america give credit cards to people like me & credit\_card\_inquiry & new\_card\\ \hline
can you pair with my phone & device\_pairing & sync\_device\\ \hline
my phone needs to be unsynced now & device\_pairing & sync\_device\\ \hline
disconnect from my phone please & device\_pairing & sync\_device\\ \hline
you need to connect to my phone & device\_pairing & sync\_device\\ \hline
when should i change my oil & schedule\>maintenance & oil\>change\>when\\ \hline
when should i get my next oil change & schedule\>maintenance & oil\>change\>when\\ \hline
when is the best time for my next oil change & schedule\>maintenance & oil\>change\>when\\ \hline
pay the cable bill with my visa card & bill\>inquiry & pay\>bill\\ \hline
my electric bill should be paid today & bill\>inquiry & pay\>bill\\ \hline
i need to pay my water bill & bill\>payment & pay\>bill\\ \hline
pay my cable bill from my checking account & bill\>payment & pay\>bill\\ \hline
pay my insurance bill & bill\>payment & pay\>bill\\ \hline
i think someone stole my card and used it & report\>fraud & report\>fraud\\ \hline
i have to report fraudulent activity on my bank of the west card & report\>fraud & report\>fraud\\ \hline
i need to tell you about my lost card & report\>lost\>card & report\>lost\>card\\ \hline
i need to report my stolen card & report\>lost\>card & report\>lost\>card\\ \hline
my capital one credit card was stolen & report\>lost\>card & report\>lost\>card\\ \hline
i require a car maintenance & schedule\>maintenance & schedule\>maintenance\\ \hline
i got to schedule some car maintenance & schedule\>maintenance & schedule\>maintenance\\ \hline
add a new meeting with tom for 6pm & meeting\>schedule & schedule\>meeting\\ \hline
would you schedule a meeting with carrie and lisa please & meeting\>schedule & schedule\>meeting\\ \hline
\end{tabular}

}
\caption{Intents generated by IntentGPT using GPT-4 and 50 shots in CLINC test set.}
\label{tab:qualitative-examples}
\end{table*}

\onecolumn
\begin{lstlisting}[breaklines=true,caption={Prompt generated by GPT-4 ICPG on MTOP Spanish}]
PROMPT: Para realizar la tarea de deteccion y descubrimiento de intenciones, debes analizar cada enunciado textual y asignarlo a una intencion especifica. Algunas de estas intenciones ya estan predefinidas, mientras que otras no y tendras que crearlas. Debes tener en cuenta las intenciones conocidas y reutilizarlas tanto como sea posible, pero tambien debes estar dispuesto a crear nuevas intenciones cuando no existan intenciones conocidas que se ajusten al enunciado dado. 

Nunca asignes un enunciado a desconocido. Por ejemplo, si el enunciado es "comparte share psychedelics y social justice con emery", la intencion seria IN:SHARE_EVENT.  Si el enunciado es puedes contestar la llamada de austin pate?, la intencion seria IN:ANSWER_CALL. Es importante que adquieras suficiente contexto y conocimiento sobre el problema y el dominio de datos especifico. Como experto en procesamiento de lenguaje natural, debes ser capaz de interpretar el significado subyacente de cada enunciado y asignarlo a la intencion correcta. 

Recuerda que el conjunto de datos puede estar en diferentes idiomas ademas del ingles. Por lo tanto, debes responder con un aviso en el mismo idioma que los ejemplos de contexto.
\end{lstlisting}
\begin{lstlisting}[breaklines=true,caption={Prompt generated by GPT-4 ICPG on MTOP French}]
PROMPT: Pour chaque phrase ou texte donne, votre tache est de determiner l'intention qui se cache derriere. L'intention est generalement liee a une action que l'utilisateur souhaite accomplir ou a une information qu'il cherche a obtenir. Par exemple, si l'utilisateur dit "repetez cette chanson encore une fois", l'intention serait "IN:REPLAY_MUSIC", indiquant que l'utilisateur souhaite rejouer la chanson actuelle. nnIl est important de noter que certaines intentions peuvent etre reutilisees pour differentes phrases. 

Par exemple, "repetez cette chanson encore une fois" et "repassez la chanson actuelle" ont tous deux l'intention "IN:REPLAY_MUSIC". Cependant, si une phrase ne correspond a aucune intention connue, vous devrez creer une nouvelle intention. Il est egalement crucial de ne jamais attribuer une phrase a l'intention 'inconnue'. Chaque phrase a une intention, meme si elle n'est pas immediatement evidente. Lors de l'attribution des intentions, veillez a prendre en compte le contexte de la phrase et a utiliser vos connaissances sur le domaine specifique des donnees. Par exemple, si l'utilisateur dit "ou se deroulent la plupart de mes rappels?", l'intention serait "IN:GET_REMINDER_LOCATION", indiquant que l'utilisateur cherche a savoir ou se deroulent la plupart de ses rappels. 

Enfin, n'oubliez pas que le format de la reponse n'est pas specifie. Vous pouvez donc repondre de la maniere qui vous semble la plus appropriee, tant que vous identifiez correctement l'intention.
\end{lstlisting}
\begin{lstlisting}[breaklines=true,caption={A. Prompt used for In-Context Prompt Generation (ICPG) on GPT-4}]
You are a helpful assistant and an expert in natural language processing. You specialize in intent detection and discovery, the task of assigning textual utterances to specific intents, some of which are pre-defined and others are not and have to be created. Keep in mind that in the problem of intent discovery, you need to be aware of the known intents and reuse them as much as possible, but need to create new intents when there are not known intents that fit the given utterance, and never assign a utterance to 'unknown'. You will be presented with a set of examples from the dataset, and need to acquire sufficient context and knowledge about the problem and specific data domain. 

As an expert, provide a detailed prompt for an AI language model to solve the task of intent discovery, maximizing the model's performance in the task. Provide effective guidelines about how to solve the task in the prompt, but keep it very concise.

EXAMPLES: {train_examples}
You must respond using the following format: 
PROMPT: <prompt>
\end{lstlisting}
\begin{lstlisting}[breaklines=true,caption={B. Generated Prompt by GPT-4 as ICPG on CLINC (KIR=0.75)}]
AI language model, your task is to assign the correct intent to a given textual utterance. The intent can be one of the pre-defined intents or a new one that you create based on the context and knowledge about the problem and specific data domain. You should never assign an utterance to 'unknown'. 

For each utterance, analyze the context and the specific request or action implied. If the utterance matches a known intent, assign it to that intent. If it doesn't match any known intent, create a new intent that accurately represents the request or action implied by the utterance.

Remember, the goal is to understand the user's intent as accurately as possible. Be aware of the known intents and reuse them as much as possible, but don't hesitate to create new intents when necessary. 

Here are some examples:

- If the utterance is "how are you named", the intent is "what_is_your_name".
- If the utterance is "i need to know the total calories for a chicken caesar salad", the intent is "calories".
- If the utterance is "please pause my banking actions", the intent is "freeze_account".

Use these examples as a guide, but remember that the utterances can vary greatly in structure and content. Your task is to understand the underlying intent, regardless of how the utterance is phrased.
\end{lstlisting}
\begin{lstlisting}[breaklines=true,caption={C. Generated Prompt by GPT-4 as ICPG on BANKING (KIR=0.75)}]

As an AI language model, your task is to assign the correct intent to a given textual utterance. The intents are related to various banking and financial services scenarios. You should be aware of the known intents and reuse them as much as possible. However, if there is no known intent that fits the given utterance, you should create a new intent. Remember, you should never assign an utterance to 'unknown'. 

The utterances can be questions, statements, or requests related to banking services like card transactions, account top-ups, refunds, identity verification, card delivery, transfer fees, and more. Your task is to understand the context and the specific intent behind each utterance. 

For example, if the utterance is "what happened to the money after i put in the wrong info and it got declined", the intent should be "topping_up_by_card". If the utterance is "i cannot see a refund in my account", the intent should be "Refund_not_showing_up". 

Please note that the same utterance can have different intents based on the context. For instance, "my card is expiring, how do i get a new one?" has the intent "card_about_to_expire", while "i have enough money in my account, so why is my card being declined?" has the intent "declined_card_payment".

Your responses should be concise, accurate, and contextually relevant.
\end{lstlisting}
\begin{lstlisting}[breaklines=true,caption={E. Basic human-generated task description}]
You are a helpful assistant and an expert in natural language processing and specialize in the task of intent detection. Your task is to assign utterances to their corresponding intent. The intent can be one of the pre-defined intents or a new one that you create based on the context and knowledge about the problem and specific data domain. You should never assign an utterance to 'unknown'. For each utterance, analyze the context and the specific request or action implied. If the utterance matches a known intent, assign it to that intent. If it doesn't match any known intent, create a new intent that accurately represents the request or action implied by the utterance. Remember, the goal is to understand the user's intent as accurately as possible. Be aware of the known intents and reuse them as much as possible, but don't hesitate to create new intents when necessary.
\end{lstlisting}
\begin{lstlisting}[breaklines=true,caption={D. Complete interaction with IntentGPT Intent Predictor on CLINC}]
You are a helpful assistant and an expert in natural language processing and specialize in the task of intent detection.
AI language model, your task is to assign the correct intent to a given textual utterance. The intent can be one of the pre-defined intents or a new one that you create based on the context and knowledge about the problem and specific data domain. You should never assign an utterance to 'unknown'. 

For each utterance, analyze the context and the specific request or action implied. If the utterance matches a known intent, assign it to that intent. If it doesn't match any known intent, create a new intent that accurately represents the request or action implied by the utterance.

Remember, the goal is to understand the user's intent as accurately as possible. Be aware of the known intents and reuse them as much as possible, but don't hesitate to create new intents when necessary. 

Here are some examples:

- If the utterance is "how are you named", the intent is "what_is_your_name".
- If the utterance is "i need to know the total calories for a chicken caesar salad", the intent is "calories".
- If the utterance is "please pause my banking actions", the intent is "freeze_account".

Use these examples as a guide, but remember that the utterances can vary greatly in structure and content. Your task is to understand the underlying intent, regardless of how the utterance is phrased.
Make sure each intent is only between one and three words, and as short and reusable as possible. Use the same format as the context examples. Don't classify the examples below CONTEXT EXAMPLES. Only classify the test examples below TEST EXAMPLES. You are prohibited to assign intents to 'unknown'. Instead, create a new intent. 

Don't discover a new intent if you have already discovered one that is similar. Make sure that the intents are not very generic, you can be fine-grained. Use the following list of known intents to keep reference, reuse them as much as possible: 

no, what_is_your_name, calories, shopping_list, freeze_account, pto_request_status, current_location, where_are_you_from, income, gas, confirm_reservation, maybe, improve_credit_score, book_hotel, repeat, apr, damaged_card, tire_pressure, balance, share_location, what_are_your_hobbies, insurance_change, car_rental, smart_home, gas_type, yes, pto_used, replacement_card_duration, order_status, cancel, restaurant_suggestion, rollover_401k, change_accent, redeem_rewards, credit_score, reminder, restaurant_reviews, meeting_schedule, meal_suggestion, exchange_rate, directions, flight_status, calendar, do_you_have_pets, alarm, travel_suggestion, update_playlist, ingredients_list, travel_notification, what_can_i_ask_you, w2, report_lost_card, book_flight, distance, thank_you, travel_alert, calculator, make_call, roll_dice, pto_balance, how_old_are_you, international_visa, how_busy, time, are_you_a_bot, timezone, change_user_name, mpg, insurance, payday, vaccines, fun_fact, report_fraud, pto_request, taxes, restaurant_reservation, measurement_conversion, last_maintenance, play_music, application_status, credit_limit_change, change_speed, date, who_made_you, pin_change, spending_history, definition, reminder_update, change_ai_name, tire_change, order, account_blocked, calendar_update, routing, cook_time, food_last, interest_rate, greeting, user_name, todo_list, ingredient_substitution, schedule_maintenance, shopping_list_update, transactions, rewards_balance, credit_limit, carry_on, expiration_date, change_language, text, next_holiday, who_do_you_work_for

CONTEXT EXAMPLES:

Utterance: please remind me later, Intent: reminder_update
Utterance: i want to know what health plan i'm currently on, Intent: insurance
Utterance: do you work for another individual, Intent: who_do_you_work_for
Utterance: tell me my health plan, Intent: insurance
Utterance: is there an insurance plan for my health, Intent: insurance
Utterance: are you working for another person or entity, Intent: who_do_you_work_for
Utterance: i'm ready to put in the order for everything on my shopping list, Intent: order
Utterance: get everything on my shopping list, Intent: order
Utterance: i'd like to you order everything that's on my shopping list, Intent: order
Utterance: it is damaging to my credit score if i miss one credit card payment, Intent: improve_credit_score

RESPONSE FORMAT:

ID: <i>, Utterance: <content>, Intent: <intent> Use the same ID in the test example.

TEST EXAMPLES (to predict):

ID:0, is there another person you work for
ID:1, does tartine in san francisco take reservations
ID:2, what is the traffic like on the road to downtown dallas
ID:3, what do you think the meaning of life is
ID:4, what is the square root of 80
ID:5, is the friday's full after 4
ID:6, i'm going to need an uber for 4 people to go to patty's bar
ID:7, place an order for my entire shopping list
ID:8, what is the traffic like on the way to town
ID:9, please spell doctor
ID:10, tell my bank i'm in canada the entire week of the 6th to the 13th
ID:11, i need help with some fun tourist things to do in tokyo
ID:12, please remind me at a later time
ID:13, what specific health plan do i have
ID:14, can i harm my credit if i miss a car payment
ID:15, please call mark

Your response:

ID:0, Utterance: is there another person you work for, Intent: who_do_you_work_for
ID:1, Utterance: does tartine in san francisco take reservations, Intent: restaurant_reservation
ID:2, Utterance: what is the traffic like on the road to downtown dallas, Intent: traffic_status
ID:3, Utterance: what do you think the meaning of life is, Intent: philosophical_question
ID:4, Utterance: what is the square root of 80, Intent: calculator
ID:5, Utterance: is the friday's full after 4, Intent: restaurant_busy_status
ID:6, Utterance: i'm going to need an uber for 4 people to go to patty's bar, Intent: book_ride
ID:7, Utterance: place an order for my entire shopping list, Intent: order
ID:8, Utterance: what is the traffic like on the way to town, Intent: traffic_status
ID:9, Utterance: please spell doctor, Intent: spelling
ID:10, Utterance: tell my bank i'm in canada the entire week of the 6th to the 13th, Intent: travel_notification
ID:11, Utterance: i need help with some fun tourist things to do in tokyo, Intent: travel_suggestion
ID:12, Utterance: please remind me at a later time, Intent: reminder_update
ID:13, Utterance: what specific health plan do i have, Intent: insurance
ID:14, Utterance: can i harm my credit if i miss a car payment, Intent: improve_credit_score
ID:15, Utterance: please call mark, Intent: make_call
\end{lstlisting}
\end{document}